\documentclass[runningheads]{llncs}

\usepackage{eccv}

\usepackage{eccvabbrv}

\usepackage{graphicx}
\usepackage{booktabs}

\usepackage[accsupp]{axessibility}  

\usepackage{wrapfig}

\usepackage{colortbl}
\usepackage{multirow}

\usepackage{hyperref}

\usepackage{orcidlink}

\begin{document}

\title{Layout-Guided Controllable Pathology Image Generation with In-Context Diffusion Transformers} 

\titlerunning{Abbreviated paper title}

\author{Yuntao Shou\inst{1} \and
Xiangyong Cao\inst{1} \and
Qian Zhao\inst{1} \and
Deyu Meng\inst{1}}

\authorrunning{Y. Shou et al.}

\institute{Xi'an Jiaotong University \\
\email{shouyuntao@stu.xjtu.edu.cn}\\
}

\maketitle

\begin{abstract}
 Controllable pathology image synthesis requires reliable regulation of spatial layout, tissue morphology, and semantic detail. However, existing text-guided diffusion models offer only coarse global control and lack the ability to enforce fine-grained structural constraints. Progress is further limited by the absence of large datasets that pair patch-level spatial layouts with detailed diagnostic descriptions, since generating such annotations for gigapixel whole-slide images is prohibitively time-consuming for human experts. To overcome these challenges, we first develop a scalable multi-agent LVLM annotation framework that integrates image description, diagnostic step extraction, and automatic quality judgment into a coordinated pipeline, and we evaluate the reliability of the system through a human verification process. This framework enables efficient construction of fine-grained and clinically aligned supervision at scale. Building on the curated data, we propose In-Context Diffusion Transformer (IC-DiT), a layout-aware generative model that incorporates spatial layouts, textual descriptions, and visual embeddings into a unified diffusion transformer. Through hierarchical multimodal attention, IC-DiT maintains global semantic coherence while accurately preserving structural and morphological details. Extensive experiments on five histopathology datasets show that IC-DiT achieves higher fidelity, stronger spatial controllability, and better diagnostic consistency than existing methods. In addition, the generated images serve as effective data augmentation resources for downstream tasks such as cancer classification and survival analysis.
\end{abstract}

\section{Introduction}
\label{sec:intro}

\begin{wrapfigure}{r}{0.5\linewidth}
    \vspace{-7mm}
    \centering
    \includegraphics[width=\linewidth]{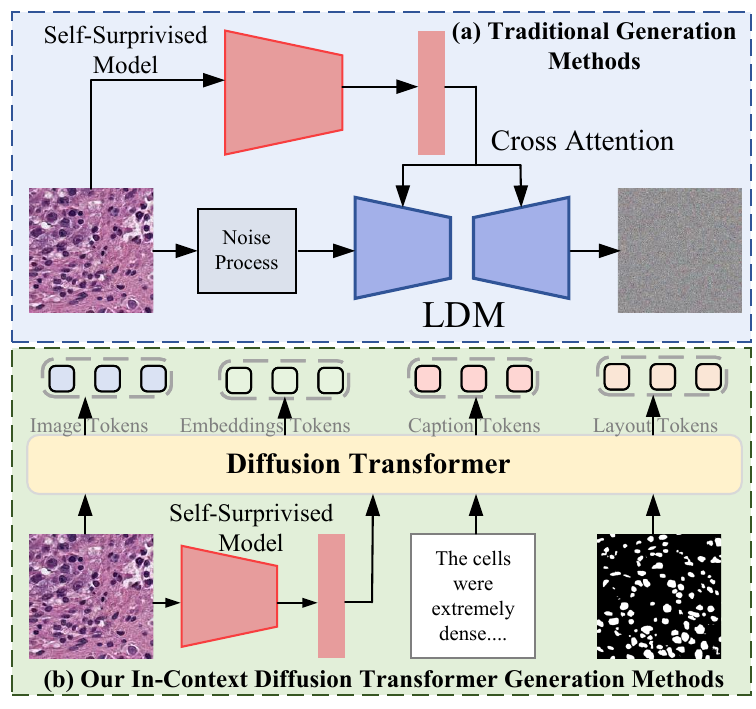}
    \vspace{-5mm}
    \caption{Comparison between traditional diffusion-based pathology synthesis and the proposed IC-DiT.
    (a) Conventional latent diffusion models rely on global cross-attention, offering limited control over spatial structure and semantics.
    (b) IC-DiT integrates multi-modal in-context conditioning with image, layout, visual, and text tokens, enabling anatomically coherent and spatially controllable pathology generation.}
    \label{fig:placeholder}
    \vspace{-8mm}
\end{wrapfigure}

The synthesis of realistic and clinically reliable pathology images remains a challenging problem in computational pathology. Although recent text-guided diffusion models have demonstrated remarkable capability in generating photorealistic images with semantic control \cite{zhang2024diffboost, xing2024svgdreamer, feng2024enhancing, lee2025text}, they still struggle to provide fine-grained spatial and morphological controllability. As shown in Fig. \ref{fig:placeholder}, most existing diffusion pipelines rely on coarse global conditions such as prompts or latent embeddings, which offer only implicit control over spatial arrangements. However, histopathological images critically depend on precise cellular organization, tissue topology, and morphological continuity, making layout-controllable generation essential for producing diagnostically meaningful outputs \cite{yellapragada2025zoomldm, ho2025f2fldm}. Without explicit structural guidance, standard diffusion models often fail to maintain consistent glandular boundaries, cell distributions, or regional tissue compositions, limiting their applicability in clinical and educational scenarios \cite{yellapragada2024pathldm}.

To address the above limitations, we aim to propose a context-aware generative framework that integrates explicit spatial layout conditioning with textual and visual semantics. However, achieving such controllable synthesis presents two fundamental challenges. First, realizing layout-aware generation requires a training dataset that contains fine-grained spatial layouts paired with detailed diagnostic descriptions, yet no publicly available pathology dataset provides such patch-level semantic layouts together with high-quality captions. Second, existing diffusion-based pathology generators lack mechanisms for precise structural supervision, making it difficult for the model to learn how spatial constraints should be mapped to valid tissue morphology. Therefore, we devote ourselves to addressing these two challenges in this work.

For constructing the desired dataset, unlike natural images that can be annotated with coarse captions by crowdworkers, histopathology data involve gigapixel whole-slide images with highly complex tissue morphology and subtle cellular differences \cite{lu2024multimodal, sun2024pathasst, shou2025graph}. Producing consistent patch-level annotations requires medical expertise, and the annotation burden grows exponentially with image scale. For example, generating high-quality descriptions for the entire BRCA cohort would require more than 40,000 hours of expert pathologists’ time \cite{graikos2024learned}. Such heavy cost makes large-scale fine-grained annotation essentially infeasible and leaves current models without the conditional signals needed for controllable synthesis \cite{chen2024wsicaption, harb2024diffusion, ahmed2024pathalign}. To overcome this bottleneck, we introduce an automated multi-agent large vision–language model (LVLM) annotation framework, enabling scalable generation of reliable pseudo-supervision for fine-grained pathology data. The framework orchestrates several specialized agents to mimic diagnostic reasoning. Specifically, Image-to-Text agent extracts morphological characteristics, a Key-Step Extraction agent decomposes them into interpretable diagnostic steps, and a Judge agent evaluates their consistency through automatic metrics. A human verification process then calibrates the scoring criteria, progressively improving domain alignment \cite{zhang2025mme, liu2025survey, chen2024we, shou2025multimodal}. This multi-agent pipeline allows us to automatically construct fine-grained textual annotations at scale, significantly reducing expert workload while maintaining diagnostic relevance.

Based on the curated dataset, we introduce In-Context Diffusion Transformer (IC-DiT), a unified generation framework that supports explicit layout-to-pathology synthesis. IC-DiT injects multimodal contextual information, including spatial layout masks, textual descriptions, and visual embeddings, into a diffusion transformer backbone. Within IC-DiT, a set of multimodal attention blocks performs hierarchical feature alignment in which textual semantics guide global appearance, layout tokens ensure structural fidelity, and visual embeddings enhance local morphological detail. This design enables controllable synthesis with fine-grained spatial accuracy and diagnostically faithful textures, addressing key limitations of previous diffusion-based approaches. The main contributions of this paper are summarized as follows.

\begin{itemize}

    \item To the best of our knowledge, we make the first attempt to incorporate large vision–language models (LVLMs) into a multi-agent annotation framework for histopathology, enabling scalable generation of fine-grained and clinically interpretable pseudo-labels through automated reasoning and human verification.

    \item We propose a unified In-Context Diffusion Transformer (IC-DiT) that integrates textual, spatial, and visual embedding modalities via dedicated encoders and hybrid cross-modal attention blocks, achieving fine-grained controllability over both semantic appearance and structural layout during pathology image synthesis.

    \item We conduct comprehensive experiments on multiple histopathology datasets, demonstrating that our method can obtain superior generation performance over state-of-the-art baselines in image fidelity, spatial consistency, and controllability. Additionally, our method can help augment the data for downstream high-level tasks, such as survival analysis and cancer classification.
\end{itemize}

\section{Related Work}

\subsection{Diffusion Models for Digital Pathology}

Diffusion models have recently emerged as powerful generative tools for high-fidelity image synthesis and data augmentation in medical imaging, including digital pathology \cite{dhariwal2021diffusion, fei2023generative, ho2020denoising, hou2023global, nichol2021improved}. By modeling the data distribution through a learned denoising process, these models can generate realistic histopathological images, aiding in data scarcity and domain generalization issues. Yellapragada et al. \cite{yellapragada2024pathldm} demonstrated the effectiveness of conditional diffusion models in synthesizing diverse and class-specific histology patches for cancer classification tasks. Similarly, Xu et al. \cite{xu2024histo} applied latent diffusion to perform high-resolution Whole Slide Image (WSI) synthesis, improving downstream segmentation and detection performance. 



\subsection{Diffusion Transformer}

Diffusion Transformer (DiT) \cite{peebles2023scalable, esser2024scaling, lv2025rethinking, shin2025exploring, ma2025x2i} represents a recent advancement in generative modeling, combining the strengths of diffusion models and vision transformers. Unlike traditional convolution-based denoising networks, DiT replaces the U-Net backbone with a pure transformer architecture, offering greater modeling capacity and scalability for high-resolution image generation. DiT operates directly in the latent space using transformer blocks, and demonstrated state-of-the-art performance on ImageNet. Its self-attention mechanism enables efficient global context modeling, which is particularly beneficial for visual domains where spatial dependencies are critical, such as digital pathology. Compared to latent diffusion models that rely on autoencoders and operate in compressed latent spaces, DiT preserves full spatial fidelity and shows superior performance in capturing fine-grained textures and structural \cite{hatamizadeh2024diffit, xu2023vit}. 



\subsection{Layout-to-Image Generation}

Layout-to-image generation creates photorealistic images from structured scene layouts, typically using bounding boxes, object categories, and spatial relationships \cite{11180796, tang2024crs, Zhang_2023_ICCV}. Early methods used conditional generative adversarial networks (cGANs)~\cite{isola2017image} or variational autoencoders (VAEs)~\cite{johnson2018image}, but struggled with multi-object composition and consistency. Later, object-aware architectures like LayoutGAN~\cite{lilayoutgan} and InstaGAN~\cite{moinstagan} improved by modeling individual object features before combining them into a scene. Transformer-based models, such as DALL·E~\cite{ramesh2021zero}, GLIDE~\cite{nichol2022glide}, LAMIC~\cite{chen2025lamic}, and LayoutDM~\cite{chai2023layoutdm}, have advanced by treating layout elements as tokens and using autoregressive or diffusion-based methods for high-fidelity synthesis. Additionally, scene graphs~\cite{johnson2018image, ashual2019specifying} and hierarchical layouts~\cite{hong2018inferring} have been explored to capture complex relational semantics beyond basic bounding boxes. Recent works have adapted layout-guided generation for explicit structural grounding to mitigate geometric hallucinations. For example, PathDiff~\cite{Bhosale_2025_ICCV} employs unpaired text and mask conditions for spatially controlled histopathology synthesis. Similarly, other recent frameworks integrate fine-grained layout guidance to generate realistic clinical scans that strictly conform to predefined anatomical structures~\cite{min2024co, oh2023diffmix, yellapragada2025pixcell}.

\section{Proposed Method}

In this section, we first briefly overview the proposed multi-agent and IC-DiT framework, and then describe the details of its components, as well as the training strategy.

\begin{figure*}
    \centering
    \includegraphics[width=0.92\textwidth]{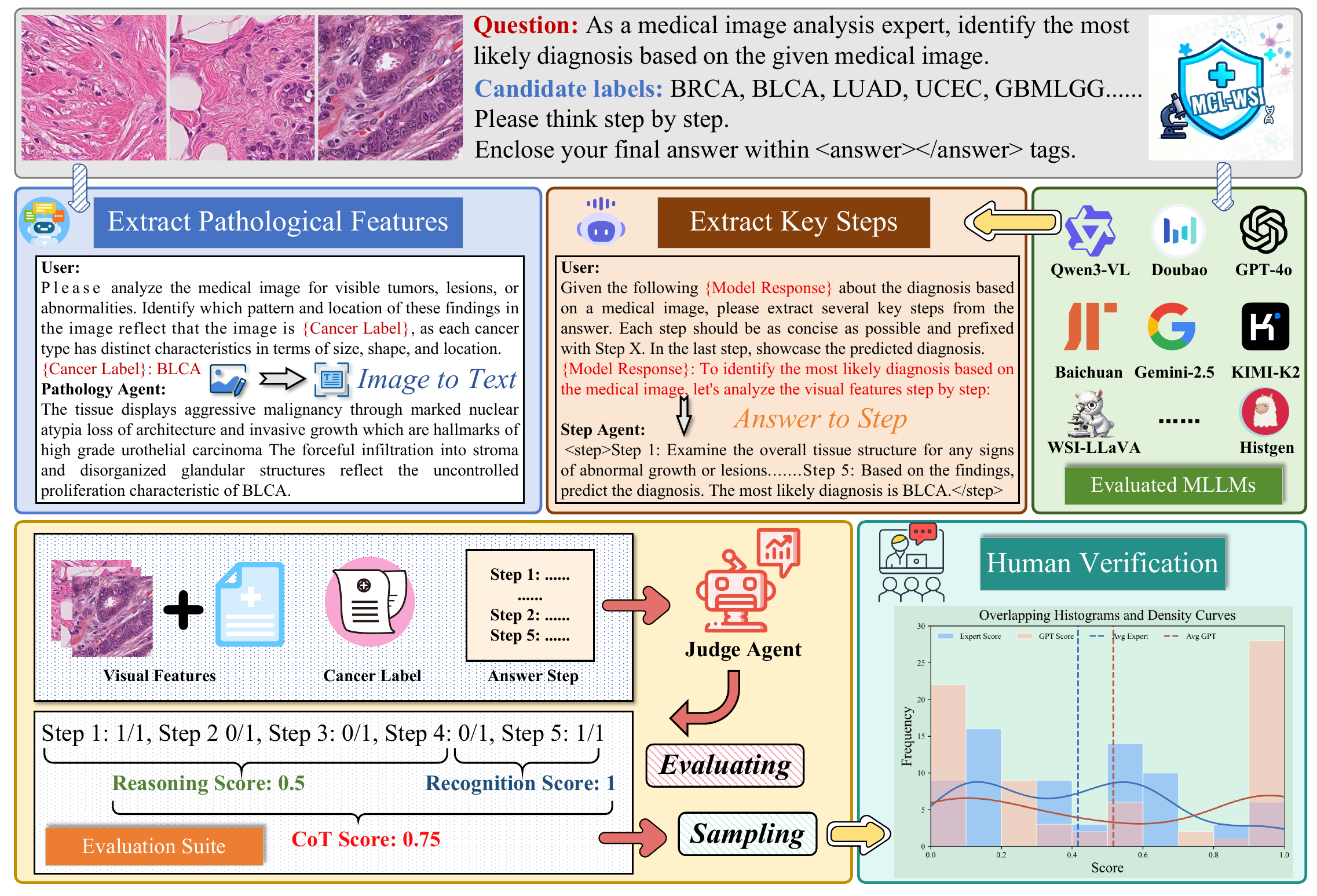}
    \caption{Overview of the multi-agent framework for pathological image description generation. It includes: (1) Pathological Feature Extraction, where vision-language models convert images into visual-textual descriptions; (2) Key Step Extraction, where language models break down diagnostic reasoning into structured steps; and (3) Human Verification, which uses automated scores and expert feedback for accuracy assessment. A Judge Agent quantifies reliability through recognition and coherence metrics.}
    \label{teaser}
    \vspace{-2mm}
\end{figure*}

\subsection{Framework Overview}

\begin{figure*}
    \centering
    \includegraphics[width=0.9\linewidth]{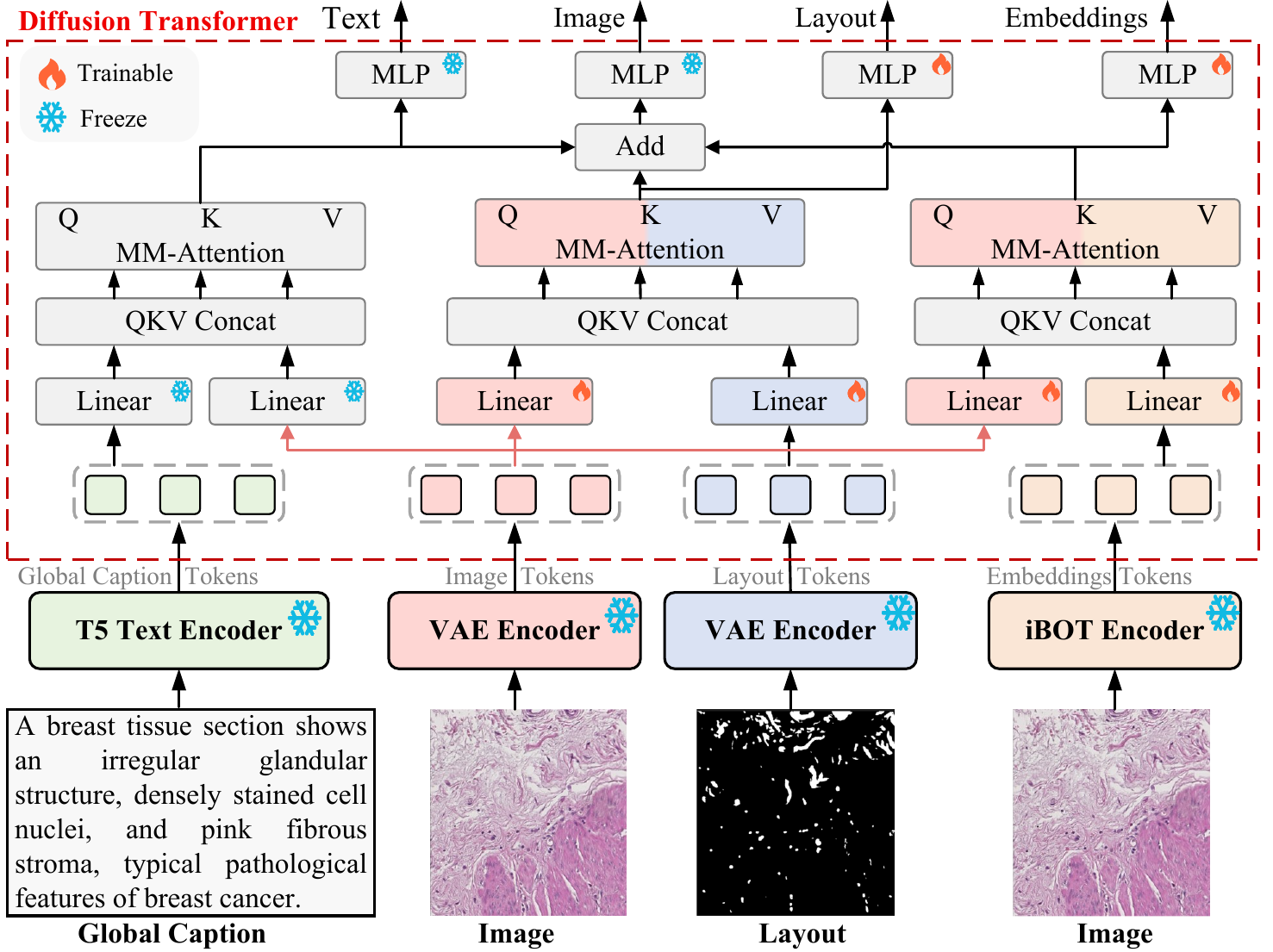}
    \vspace{-2mm}
    \caption{Overview of the In-Context Diffusion Transformer (IC-DiT) architecture. This model integrates text, image, layout, and embedding tokens through separate encoders (T5, VAE, iBOT) and processes them with multi-modal attention (MM-Attention) mechanisms. The attention mechanism is applied across different modalities to fuse them in a shared latent space, allowing for high-fidelity pathology image generation based on semantic descriptions and spatial layouts.}
    \vspace{-2mm}
    \label{fig:His_DiT}
\end{figure*}

Figs.~\ref{teaser} and~\ref{fig:His_DiT} show our framework, combining a multi-agent annotation pipeline with In-Context Diffusion Transformer (IC-DiT). The multi-agent system includes: (1) Pathological Feature Extraction, using LVLMs to convert images into visual-textual descriptions; (2) Step-wise Diagnostic Decomposition, breaking down clinical reasoning into structured steps; and (3) Human Verification, integrating automated scoring and expert feedback for accuracy. IC-DiT consists of: (1) a T5 encoder for semantic prompts; (2) dual frozen VAE encoders for image and layout features; (3) a self-supervised iBOT encoder for anatomical priors; and (4) a diffusion transformer with hybrid multimodal attention. Modality features are mapped into a shared space, and MM-Attention enables fine control over both semantic and spatial structure during generation.

\subsection{Pathology Image-Text Pair Construction}

To enable controllable diffusion-based pathology image generation with fine-grained supervision, we propose a multi-agent annotation framework that constructs semantically consistent patch-level image–text pairs. Each local tissue region is independently analyzed, described, and verified through coordinated LVLM reasoning and human evaluation as shown in Fig. \ref{teaser}. In the following, we discuss the detail of the multi-agent annotation framework. 

\textbf{Step-Wise Diagnostic Reasoning.} Formally, a whole-slide image $I \in \mathbb{R}^{H\times W \times 3}$ is divided into a set of $N$ non-overlapping patches as  follows:
\begin{equation}
I = \{x_1, x_2, \ldots, x_N\}, \quad x_i \in \mathbb{R}^{h\times w\times 3}.
\end{equation} 

A reasoning agent $g_{\text{step}}$ first performs localized observation and deduction to form a fine-grained diagnostic chain as follows:
\begin{equation}
Y_i=\{y_{i,1},y_{i,2},\dots,y_{i,M_i}\}, \quad y_{i,j}=g_{\text{step}}(x_i,\text{prompt}),
\end{equation}
where each $y_{i,j}$ is a structured statement such as
``dense epithelial clustering'' or ``nuclei show hyperchromatic features''.
The sequence $Y_i$ defines a directed reasoning graph
$\mathcal{G}_i=(\mathcal{V}_i,\mathcal{E}_i)$ with $\mathcal{V}_i=\{y_{i,j}\}$ and
$\mathcal{E}_i=\{(y_{i,j},y_{i,k})|j<k\}$.
A lightweight graph aggregator $\Psi$ yields a probabilistic patch label distribution:
\begin{equation}
\mathbf{p}_i(\ell)=\text{Softmax}\!\left(\Psi(\{y_{i,j}\}_{j=1}^{M_i})\right), \quad
\hat{\ell}_i=\arg\max_{\ell\in\mathcal{C}} \mathbf{p}_i(\ell),
\end{equation}
where $\mathcal{C}$ denotes the pathology label set.

\textbf{Label-Conditioned Patch-Level Visual-Language Description.}
The inferred label $\hat{\ell}_i$ is then injected as conditional context into a
label-aware LVLM to produce a clinically coherent description:
\begin{equation}
t_i=f_{\text{LVLM}}\!\big(x_i,\text{prompt}(\hat{\ell}_i)\big),
\end{equation}
yielding a localized caption such as ``fibrotic stroma with mild lymphocytic infiltration''.

\textbf{Judge Agent for Global Evaluation.}
To evaluate the reliability and quality of each automatically annotated patch, 
we employ a powerful large vision--language model (LVLM) as a judgment agent. 
This agent directly assesses the overall correctness, diagnostic consistency, and textual coherence of the generated sample, 
serving as an automatic evaluator that mimics expert reasoning behavior. 
Given a patch $x_i$ with its reasoning chain $Y_i$, inferred label $\hat{\ell}_i$, 
and generated textual description $t_i$, the judgment agent $J(\cdot)$ is prompted with a structured query $P_J$ 
that requests an analytical evaluation of the sample across multiple diagnostic perspectives. 
The evaluation process can be formulated as:
\begin{equation}
Q_i = J(P_J,\, x_i,\, Y_i,\, \hat{\ell}_i,\, t_i),
\end{equation}
where $Q_i \in [0,1]$ denotes the confidence score provided by the LVLM. 
The score is directly derived from the model’s textual output via instruction-based prompting, 
in which the LVLM is asked to rate the sample quality along several explicit dimensions 
(\emph{e.g.}, visual grounding accuracy, reasoning validity, and factual consistency) 
and then summarize them into an overall confidence score.

\textbf{Human Verification.} To further validate the reliability of our multi-agent annotation and evaluation framework, we conducted a comprehensive human verification study with multiple pathologists. Each expert independently reviewed a subset of patch-level samples drawn from the automatically annotated corpus, which included the visual patch $x_i$, its reasoning chain $Y_i$, the predicted label $\hat{\ell}_i$, and the generated textual description $t_i$. The experts were asked to assess these samples in terms of visual grounding accuracy, diagnostic reasoning soundness, and textual fidelity, following standard pathological evaluation criteria. To ensure consistency, all experts followed a unified scoring guideline adapted from clinical diagnostic reporting. The consistency between human ratings and the scores produced by our LVLM-based judgment agent was then qualitatively analyzed. The results revealed a strong alignment between the model’s evaluations and expert judgments, indicating that the proposed LVLM-as-judge mechanism can effectively capture domain-relevant cues and provide assessments highly consistent with human reasoning patterns.

\subsection{Layout Representation Generation}

We extract spatial layout representations using UN-SAM \cite{chen2024sam}, a zero-shot segmentation model for nuclei and tissue structures. Given a whole-slide or patch image, UN-SAM produces binary masks that outline nuclei and stroma without task-specific fine-tuning. These masks are then encoded by a VAE into a compact latent layout aligned with the diffusion transformer. This layout serves as a structured prior, enabling anatomically precise and multi-modally compatible controllable generation.

\subsection{Layout-to-Image Generation}

\textbf{Tokenization of Multi-modal Conditions.} In our approach, we tokenize the input modalities to facilitate their integration within a unified diffusion transformer framework. For textual information, we employ a frozen T5 encoder \cite{raffel2020exploring} to process the global prompt, extracting contextual embeddings that represent the semantic guidance for image generation. Regarding the target image, it is first encoded into a latent space using a frozen VAE, from which the latent features are patchified into tokens that capture essential visual elements. Spatial structures are encoded by processing layout masks through a dedicated VAE encoder \cite{batifol2025flux}, also frozen, with the resulting latent features being patchified to generate layout tokens, thereby capturing rich spatial priors without relying on explicit coordinates or bounding boxes. To enrich the model with additional high-level context, we utilize a frozen iBOT encoder \cite{zhou2021ibot} to derive dense visual embeddings from either the raw image or reference patches, providing supplementary information that enhances the generative process. This multi-modal tokenization strategy enables coherent alignment and effective fusion of semantics, spatial structure, and appearance, driving precise and controllable pathology image synthesis.

\subsection{Semantic, Spatial, and Visual Integration}

We integrate semantic, spatial, and visual information through a unified multi-modal attention (MM-Attention) mechanism that enables cross-modal alignment in the shared latent space. As illustrated in Fig.~\ref{fig:His_DiT}, the global caption tokens $\mathbf{h}^p$, image tokens $\mathbf{h}^z$, layout tokens $\mathbf{h}^l$, and embedding tokens $\mathbf{h}^e$ are projected into query $\mathbf{Q}$, key $\mathbf{K}$, and value $\mathbf{V}$ spaces via modality-specific linear layers. These projections are then concatenated across modalities before being fed into a shared attention module. The MM-Attention operation is defined as:
\begin{equation}
\mathbf{h}^{z'}, \mathbf{h}^{p'} = \mathrm{Attention}\left( [\mathbf{Q}^z, \mathbf{Q}^p], [\mathbf{K}^z, \mathbf{K}^p], [\mathbf{V}^z, \mathbf{V}^p] \right),
\end{equation}
where $[\cdot, \cdot]$ denotes concatenation along the token dimension. This allows image and text modalities to interact bidirectionally, aligning visual content with semantic descriptions.

Similarly, layout tokens $\mathbf{h}^l$ are fused with image tokens:
\begin{equation}
\mathbf{h}^{z'}, \mathbf{h}^{l'} = \mathrm{Attention}\left( [\mathbf{Q}^z, \mathbf{Q}^l], [\mathbf{K}^z, \mathbf{K}^l], [\mathbf{V}^z, \mathbf{V}^l] \right),
\end{equation}
enabling spatial structure to guide the generation of anatomical features while preserving layout consistency.

Finally, embeddings from iBOT are incorporated via:
\begin{equation}
\mathbf{h}^{z'}, \mathbf{h}^{e'} = \mathrm{Attention}\left( [\mathbf{Q}^z, \mathbf{Q}^e], [\mathbf{K}^z, \mathbf{K}^e], [\mathbf{V}^z, \mathbf{V}^e] \right),
\end{equation}
which enhances fine-grained visual fidelity by injecting high-level contextual representations. By unifying these interactions within a single diffusion transformer, our model achieves coherent control over semantics, spatial layout, and appearance, ensuring high-fidelity pathology image generation that is both interpretable and clinically relevant.

\paragraph{Training and Inference.}
We freeze all pre-trained components, including the T5 text encoder, VAE encoders for image and layout, and the iBOT model, as well as train only the newly introduced learnable parameters indicated in Fig. \ref{fig:His_DiT}. The training objective is a layout and embedding-conditioned denoising loss as:
\begin{equation}
\mathcal{L}_{\text{cond}} = \mathbb{E}_{\mathbf{z}, \mathbf{p}, \mathbf{l}, \mathbf{e}, t, \epsilon \sim \mathcal{N}(0,1)} \left[ \left\| \epsilon - \epsilon_{\{\theta\}}(\mathbf{z}_t, t, \mathbf{p}, \mathbf{l}, \mathbf{e}) \right\|_2^2 \right],
\end{equation}
where $\mathbf{z}_t$ denotes the noisy latent at diffusion step $t$, $\mathbf{p}$ is the global caption, $\mathbf{l}$ is the layout representation, and $\mathbf{e}$ is the visual embedding extracted by iBOT from the image.

\section{Experiments}

\subsection{Datasets and Evaluation Metrics}

\textbf{Datasets.} The Cancer Genome Atlas (TCGA) is a public repository of genomic and clinical data from thousands of cancer patients across 33 cancer types, widely used in survival analysis to identify genetic and molecular markers linked to prognosis. In this work, we evaluate our model on five TCGA cohorts, including BLCA (373), BRCA (956), GBMLGG (569), LUAD (453), and UCEC (480). 

\noindent\textbf{Evaluation Metrics.} We adopt five metrics for assessing the generation quality: for visual quality, we report Fréchet Inception Distance computed on Inception-v3 features of real and generated pathology images (“Vanilla FID”); for high-level semantic alignment, we compute FID in the CLIP image embedding space between generated and ground-truth images (“CLIP FID”); for structural and textural fidelity, we measure the cosine similarity between visual embeddings of generated and original images extracted by a frozen iBOT model (“Embedding Similarity”); for image and text similarity, we report PLIP \cite{huang2023visual} cosine similarity computed on generated image and text embeddings; and for mask-to-image faithfulness, we report the faithfulness scores, defined as the Dice coefficient between the segmentation masks predicted from the synthesized images and those from their real counterparts. In downstream cancer analysis tasks, we further evaluate classification performance using accuracy and assess survival prediction using the concordance index (C-index).

\noindent\textbf{Baselines.} (1) For pathological image generation, we compare with MFDiffusion \cite{moghadam2023morphology}, Medfusion \cite{muller2023multimodal}, PathLDM \cite{yellapragada2024pathldm}, SSL-guided LDM \cite{graikos2024learned}, ZoomLDM \cite{yellapragada2025zoomldm}, and PathDiff \cite{Bhosale_2025_ICCV}. (2) For vision-language models evaluation, we compare with Qwen2-VL \cite{wang2024qwen2}, Qwen2.5-VL \cite{qwen2.5-VL}, Qwen3-VL \cite{bai2025qwen3}, Janus-Pro \cite{chen2025janus}, and MedGemma \cite{sellergren2025medgemma}. (3) For text-to-image alignment evaluation,we compare with ControlNet \cite{Zhang_2023_ICCV}, and PathDiff \cite{Bhosale_2025_ICCV}. (4) For image faithfulness evaluation, we compare with Diffmix \cite{oh2023diffmix}, SDM \cite{wang2022semantic}, ControlNet \cite{Zhang_2023_ICCV}, and PathDiff \cite{Bhosale_2025_ICCV}.
(5) For downstream cancer analysis, we use ViLa-MIL \cite{shi2024vila} and PMIL \cite{yu2023prototypical} for classification, and Porpoise \cite{chen2022pan}, HFBSurv \cite{li2022hfbsurv}, CMTA \cite{zhou2023cross}, and MurreNet \cite{zhou2023cross} for survival prediction.

\subsection{Pathological Image Generation}

\textbf{Image Quality Results.} Our method consistently outperforms existing approaches across all datasets and magnification levels as shown in Table \ref{tab:gen}. Among the baselines, earlier methods struggle to capture fine pathological details and structural coherence, while recent diffusion-based models show improved fidelity but limited layout control. In contrast, our proposed IC-DiT with layout-aware conditioning and multimodal in-context learning achieves the best balance between semantic accuracy, anatomical plausibility, and visual realism. Notably, IC-DiT maintains strong performance even at 5× magnification, demonstrating robustness across scales and highlighting the effectiveness of our unified multimodal framework.

\begin{table*}[htbp]
\vspace{-2mm}
\caption{Quantitative evaluation of pathological image generation methods under 20× and 5× magnification across four cancer datasets. The best and second best results are highlighted in \textbf{bold} and \underline{underline}, respectively.}
\vspace{-2mm}
\resizebox{\textwidth}{!}{
\begin{tabular}{cccc|ccc|ccc|ccc}
\hline
\multirow{2}{*}{Methods} & \multicolumn{3}{c|}{BLCA}                                                                            & \multicolumn{3}{c|}{BRCA}                                                                            & \multicolumn{3}{c|}{GBMLGG}                                                                          & \multicolumn{3}{c}{LUAD}                                                                            \\ \cline{2-13} 
                         & \multicolumn{1}{c}{Vanilla FID} & \multicolumn{1}{c}{CLIP FID} & \multicolumn{1}{c|}{Similarity} & \multicolumn{1}{c}{Vanilla FID} & \multicolumn{1}{c}{CLIP FID} & \multicolumn{1}{c|}{Similarity} & \multicolumn{1}{c}{Vanilla FID} & \multicolumn{1}{c}{CLIP FID} & \multicolumn{1}{c|}{Emb Similarity} & \multicolumn{1}{c}{Vanilla FID} & \multicolumn{1}{c}{CLIP FID} & \multicolumn{1}{c}{Similarity} \\ \hline
\textbf{\textit{20x Magnification}}     &                                 &                              &                                     &                                 &                              &                                        &                                 &                              &                                     &                                 &                              &                                    \\
MFDiffusion   &    115.73                             &  101.88                            &      0.24                               &        104.72                         &       94.31                       &          0.29                                        &       127.90                          &            115.43                  &        0.18                             &    145.72                             &     131.93                         &     0.15                               \\
Medfusion      &     57.62                             &      50.09                        &    0.33                                 &      68.52                           &    59.02                          &       0.31                              &      70.59                           &      66.33                        &       0.27                              &      73.48                           &            67.65                  &      0.24                              \\
Pathldm                  & 18.62            &    15.47      &      0.46          &  21.38        &   17.59     &   0.43            &     22.54        &   19.41      &   0.41             &    24.10         &   22.03       &   0.38             \\
SSL-guided LDM           &   10.47          &    7.43   &      0.56          &      12.57       &         8.41            &       0.49      &   14.59      &   13.36             &   0.46         &  15.03        & 14.17  & 0.44            \\
ZoomLDM         &     8.21                             &          6.85                    &  0.61                                   &           \underline{9.12}                      &       \underline{7.23}                        &     \underline{0.54}                                &     \underline{8.76}                             &        \underline{7.58}                       &              \underline{0.57}                       &    \underline{9.85}                             &    \underline{7.91}                           &     \underline{0.53}                               \\ 
PathDiff     &   \underline{7.84}   &   \underline{6.13}     &   \underline{0.65}      &   10.03     &    7.87  &    0.51        &   9.21    &   9.58   &   0.50     &    11.73     & 10.41 &  0.49\\
\rowcolor{gray!20}
IC-DiT (Ours)            &     \textbf{5.12}                            &    \textbf{5.87}                          &  \textbf{0.72}                                   &    \textbf{6.45}                            &    \textbf{6.54}                          &   \textbf{0.64}                                  &      \textbf{6.18}                           &      \textbf{6.02}                        &     \textbf{0.65}                                &    \textbf{7.23 }                             &                    \textbf{7.18 }         &        \textbf{0.62}                            \\ \hline
\textbf{\textit{5x Magnification}}                  &                                 &                              &                                     &                                 &                              &                                     &                                 &                              &                                     &                                 &                              &                                    \\
MFDiffusion      &     122.54                           &       109.29                        &    0.20                                  &   112.34                              &   100.59                           &   0.24                                  &   135.47                              &  122.1 8                            &    0.15                                  &         152.83                        &     139.40                         &      0.12                              \\
Medfusion      &      64.37                           &       56.22                        &  0.28                                   &    74.15                             &      64.83                         &      0.26                               &      77.52                           &     73.10                         &     0.22                                &     80.64                            &               74.93               &      0.19                              \\
Pathldm                  &        25.16     &    21.32      &     0.41                                &                 26.99                 &       22.84                        &     0.38                               &       28.41                          &     24.67                         &   0.36                                  &    27.58                             &      23.20                        &  0.33                                  \\
SSL-guided LDM           &      16.24                           &          12.73                    &                     0.51                 &    18.19                            &     13.84                         &     0.43                                &      17.31                           &   13.47                           &   0.47                                  &   19.05                              &             14.69                 &    0.42                                \\
ZoomLDM      &    15.37                             &    13.91                           &  0.55                                   &     16.28                            &   14.33                                                                 &      0.48                           &     15.85                         &            14.62                         &     0.51                            &    17.11                           &   15.27   &      0.47                        \\
PathDiff         &     \underline{13.84}    & \underline{12.05}  &   \underline{0.59}   &  \underline{15.11}      &    \underline{13.39 }    & \underline{0.51}  &    \underline{14.30}    &   \underline{13.29}      & \underline{0.53}  &   \underline{15.52}     &   \underline{14.26}      & \underline{0.51}\\
\rowcolor{gray!20}
IC-DiT (Ours)            &    \textbf{11.26}                             &   \textbf{11.93}                                                                &   \textbf{0.66}                              &   \textbf{12.47}                           &        \textbf{10.59}                              &   \textbf{0.58}                              &     \textbf{12.04}                          &    \textbf{11.44}                                  &      \textbf{0.59}                           &                 \textbf{13.28}             &        \textbf{13.21} & \textbf{0.56}                             \\ \hline
\end{tabular}}
\label{tab:gen}
\vspace{-2mm}
\end{table*}

\begin{wraptable}{r}{0.4\textwidth}
\vspace{-10mm}
\caption{Quantitatively evaluation the image-text similarity of four cancer datasets at 20x and 5x magnification. The best and second best results are highlighted in \textbf{bold} and \underline{underline}, respectively.}
\label{tab:simi}
\scalebox{0.7}{
\begin{tabular}{lcccc}
\hline
Methods & BLCA  & BRCA  & GBMLGG & LUAD  \\ \hline
\multicolumn{5}{l}{20x Magnification}                     \\
ControlNet               & 20.67 & 19.76 & 20.03  & 20.34 \\
PathDiff                 & \underline{22.48} & \underline{21.35} & \underline{21.92}  & \underline{22.15} \\
\rowcolor{gray!20}
IC-DiT                   & \textbf{24.83} & \textbf{24.64} & \textbf{23.72}  & \textbf{25.67} \\ \hline
\multicolumn{5}{l}{5x Magnification}                      \\
ControlNet               & 17.23 & 18.91 & 17.34  & 18.45 \\
PathDiff                 & \underline{18.34} & \underline{19.45} & \underline{18.56}  & \underline{19.59} \\
\rowcolor{gray!20}
IC-DiT                   & \textbf{20.65} & \textbf{21.29} & \textbf{21.82}  & \textbf{22.49} \\ \hline
\end{tabular}}
\vspace{-9mm}
\end{wraptable}

\noindent \textbf{Text-to-Image Alignment.} To evaluate semantic alignment, we assess image-text similarity across four cancer cohorts at $20\times$ and $5\times$ magnifications using the PLIP cosine similarity metric. As shown in Table \ref{tab:simi}, IC-DiT consistently outperforms ControlNet and PathDiff across all datasets and scales. Notably, at $20\times$ magnification, our method achieves superior scores. This performance lead persists at $5\times$ magnification, where IC-DiT sets the benchmark for cross-modal coherence. The significant margin over PathDiff indicates that our in-context mechanism facilitates more robust integration of textual prompts with spatial structures, ensuring synthesized morphology remains strictly aligned with clinical semantics.

\noindent \textbf{Vision-Language Models Evaluation.} To rigorously evaluate the quality of the generated data, we benchmarked IC-DiT across multiple cancer cohorts using various visual language models (VLMs) at 20x magnification. Our method was validated in multiple dimensions on both general and medical-specific architectures. We observed a significant scaling effect in our evaluations, with larger VLM variants consistently producing lower FID values than smaller models, indicating that IC-DiT effectively captures fine morphological nuances detectable by high-capacity discriminators. Further experimental results are provided in Section 4.2 of the supplementary materials.

\begin{wraptable}{r}{0.4\textwidth}
\vspace{-12mm}
\caption{Quantitatively evaluation the faithfulness scores of four cancer datasets at 20x and 5x magnification. The best and second best results are highlighted in \textbf{bold} and \underline{underline}, respectively.}
\label{tab:dice}
\scalebox{0.68}{
\begin{tabular}{lcccc}
\hline
Methods & BLCA  & BRCA  & GBMLGG & LUAD  \\ \hline
\multicolumn{5}{l}{20x Magnification}                     \\
Diffmix   & 0.7123   &  0.7312  & 0.7089    & 0.7245  \\
SDM   &  0.6532  & 0.6891   &  0.6245   &  0.6978 \\
ControlNet               & 0.7425 & 0.7034 & 0.7289  & 0.7167    \\
PathDiff                 & \underline{0.7856} & \underline{0.7912} & \underline{0.7734}  & \underline{0.8015} \\
\rowcolor{gray!20}
IC-DiT                   & \textbf{0.8319} & \textbf{0.8412} & \textbf{0.8352}  & \textbf{0.8363} \\ \hline
\multicolumn{5}{l}{5x Magnification}                      \\
Diffmix   &  0.6912  & 0.6945   &  0.6712   & 0.6886  \\
SDM   &  0.6104  &  0.6756  &  0.6178   &  0.6389 \\
ControlNet         & 0.6945  & 0.6889 & 0.7056  &   0.6923     \\
PathDiff                 & \underline{0.7745} & \underline{0.7801} & \underline{0.7548}  & \underline{0.7663} \\
\rowcolor{gray!20}
IC-DiT                   & \textbf{0.8245} & \textbf{0.8167} & \textbf{0.8312}  & \textbf{0.8198} \\ \hline
\end{tabular}}
\vspace{-8mm}
\end{wraptable}

\noindent \textbf{Mask-to-Image Faithfulness.} To assess the structural integrity of the generated images, we evaluate the faithfulness scores. As shown in Table \ref{tab:dice}, IC-DiT achieves the highest faithfulness across all cancer cohorts at both $20\times$ and $5\times$ magnifications, significantly outperforming competitive baselines such as PathDiff and ControlNet. Specifically, at $20\times$ magnification, our method maintains consistent structural alignment with scores exceeding 0.81 across all datasets, demonstrating its superior ability to preserve complex tissue morphology. The substantial margin over diffusion-based alternatives like SDM suggests that our layout-guided in-context mechanism effectively constrains the generative process, ensuring that the spatial arrangements of pathological structures remain anatomically accurate.

\begin{table}[htbp]
\centering
\vspace{-2mm}
\caption{The results of ablation studies in TCGA datasets. We highlight the best performing scores in \textbf{bold}.}
\vspace{-2mm}
\scalebox{0.8}{
\begin{tabular}{ccccccc}
\hline
\multicolumn{3}{c}{Components}                                & BLCA        & BRCA        & GBMLGG      & LUAD        \\ \hline
Global Caption & Layout       & Embeddings Similarity & Vanilla FID & Vanilla FID & Vanilla FID & Vanilla FID \\ \hline
$\checkmark$   &              &                              & 74.59       & 79.02       & 76.60       & 80.45       \\
               & $\checkmark$ &                              & 70.05       & 72.31       & 72.29       & 76.53       \\
               &              & $\checkmark$                 & 12.15       & 15.77       & 18.68       & 20.03       \\
$\checkmark$   & $\checkmark$ &                              & 66.14       & 67.51       & 68.20       & 73.37       \\
$\checkmark$   &              & $\checkmark$                 & 10.09       & 12.41       & 14.57       & 17.30       \\
               & $\checkmark$ & $\checkmark$                 & 7.38        & 9.52        & 10.05       & 13.47       \\
$\checkmark$   & $\checkmark$ & $\checkmark$                 & \textbf{5.12}        & \textbf{6.45} & \textbf{6.18}        & \textbf{7.23}        \\ \hline
\end{tabular}}
\vspace{-2mm}
\label{aba}
\end{table}

\noindent \textbf{Ablation Studies.} We ablate key components of IC-DiT to assess their contributions as shown in Table \ref{aba}. Using only global semantic guidance yields reasonable generation quality but lacks structural precision. Incorporating layout significantly improves anatomical fidelity by enforcing spatial consistency, while adding self-supervised embeddings further enhances pathological realism through fine-grained visual priors. The full model achieves the best performance, demonstrating that multi-modal conditioning is essential for high-quality, controllable pathology image synthesis. Notably, replacing the global caption with simple labels still provides effective semantic guidance, making the framework practical even when detailed captions are unavailable.

\begin{figure*}
    \centering
    \vspace{-2mm}
    \includegraphics[width=1\linewidth]{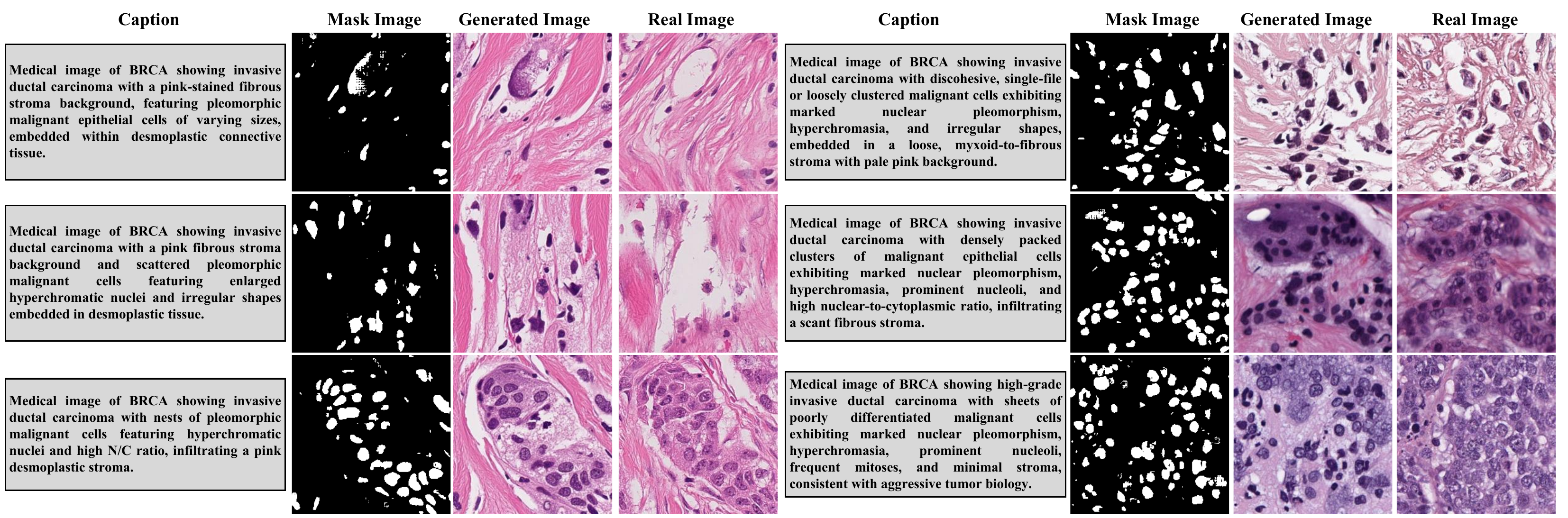}
    \vspace{-2mm}
    \caption{Qualitative comparison of layout-guided pathology image generation. Each column shows the input caption, corresponding mask, real image, and generated image.}
    \label{fig:surv_vis}
    \vspace{-2mm}
\end{figure*}

\noindent \textbf{Qualitative Visual Assessment.} As shown in Fig. \ref{fig:surv_vis}, our IC-DiT framework generates high-fidelity pathology images that faithfully reproduce both the textual semantics and spatial layouts. The generated images preserve key histological features with strong alignment to the input masks and captions. Compared to prior methods, our approach demonstrates superior layout consistency and semantic control, enabling realistic and interpretable synthesis of complex tissue structures without manual annotation.

\begin{figure*}
    \centering
    \vspace{-2mm}
    \includegraphics[width=0.95\linewidth]{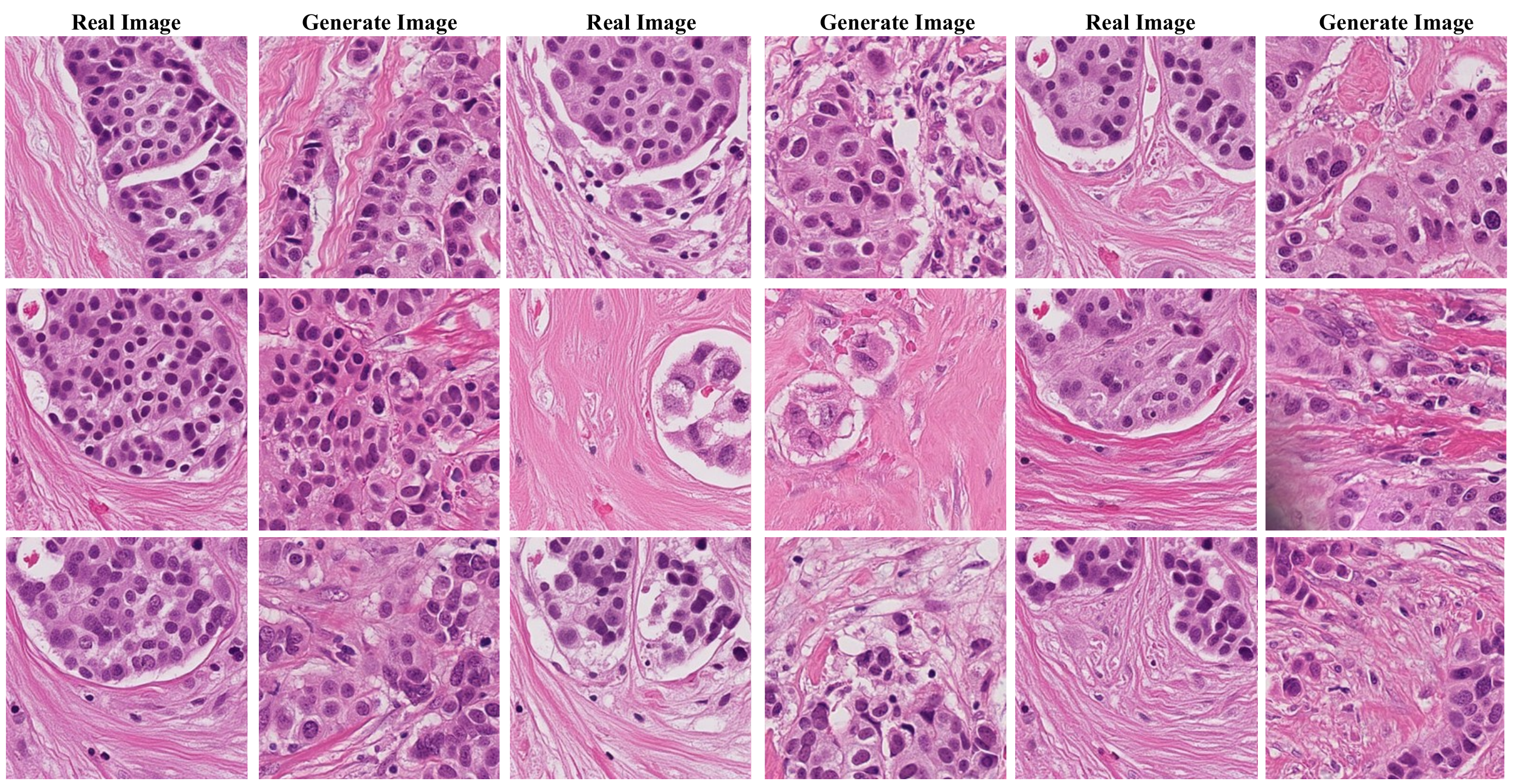}
    \caption{Ablation study on layout control. Comparison between real pathology patches and images generated without layout-mask constraints. }
    \label{fig:vis_gen}
    \vspace{-2mm}
\end{figure*}

Furthermore, removing the layout mask constraint leads to a significant degradation in local structural fidelity and tissue morphology as illustrated in Fig. \ref{fig:vis_gen}. While the model maintains coarse spatial distributions, it fails to preserve precise regional boundaries, resulting in misaligned tumor clusters and blurred tissue interfaces. This lack of explicit guidance triggers microscopic morphological drift, evidenced by inconsistent nucleus sizes and the loss of stromal fiber directionality, which suggests that unconstrained diffusion priors alone cannot sustain fine-grained pathological textures. Furthermore, text prompts prove insufficient for spatial grounding as critical structures like glandular boundaries exhibit positional shifts when explicit region allocation is absent. These observations collectively confirm that layout conditioning is indispensable for IC-DiT to achieve strict semantic-to-spatial alignment and high-fidelity synthesis.

\begin{table*}
   \caption{Comparison of survival prediction models trained with and without synthetic pathology data. C-index (mean $\pm$ std) is reported across five histopathology datasets under three training settings: (1) real data only, (2) synthetic data only and (3) real data with one-fold synthetic augmentation.
Best results are highlighted in \textbf{bold}.}
      \centering
   \vspace{-2mm}
   \small
   \scalebox{0.7}{
   \centering
      \begin{tabular}{lcccccc}
         \toprule \toprule
         {\multirow{2}{*}{Methods}}        & \multirow{2}{*}{Train Data}  & \multicolumn{5}{c}{Datasets}                                                                                                                                                                                      \\ \cmidrule(lr){3-7}
                                           &                            & BLCA                              & BRCA                              & GBMLGG                            & LUAD                              & UCEC                               \\
         \midrule
         Porpoise       &    Real                                & 0.6461 $\pm$ 0.0338               & 0.6207 $\pm$ 0.0544               & 0.8479 $\pm$ 0.0128               & 0.6403 $\pm$ 0.0412               & 0.6918 $\pm$ 0.0488 \\
        w/PathDiff    &    Synth                                           & 0.5735 $\pm$ 0.0273 & 0.5581 $\pm$ 0.0373 & 0.7383 $\pm$ 0.0470  & 0.5637 $\pm$ 0.0509  & 0.6049 $\pm$ 0.0286    \\ 
        w/PathDiff    &    Real+Synth                                         & 0.6523 $\pm$ 0.0179  & 0.6291 $\pm$ 0.0307  & 0.8585 $\pm$ 0.0288  & 0.6524 $\pm$ 0.0283  &  0.6997 $\pm$ 0.0416   \\ 
        \rowcolor{gray!15}
        w/IC-DiT    &    Synth                                           & 0.5970 $\pm$ 0.0319   & 0.5891 $\pm$ 0.0274  & 0.7613 $\pm$ 0.0482  & 0.5933 $\pm$ 0.0144  & 0.6318 $\pm$ 0.0337    \\ 
         \rowcolor{gray!15}
         w/IC-DiT     &     Real+Synth                                         & \textbf{0.6694 $\pm$ 0.0456}               & \textbf{0.6440 $\pm$ 0.0734}               & \textbf{0.8702 $\pm$ 0.0173}               & \textbf{0.6636 $\pm$ 0.0556}               & \textbf{0.7151 $\pm$ 0.0659}    \\ \midrule
         HFBSurv      &     Real                              & 0.6398 $\pm$ 0.0277               & 0.6473 $\pm$ 0.0346               & 0.8383 $\pm$ 0.0128               & 0.6501 $\pm$ 0.0495               & 0.6421 $\pm$ 0.0445                \\
        w/PathDiff    &    Synth                                           & 0.5639 $\pm$ 0.0358  & 0.5714 $\pm$ 0.0371  & 0.7418 $\pm$ 0.0474  & 0.5619 $\pm$ 0.0448  & 0.5782 $\pm$ 0.0305    \\ 
        w/PathDiff    &    Real+Synth                                         & 0.6473 $\pm$ 0.0336  & 0.6518 $\pm$ 0.0134  & 0.8452 $\pm$ 0.0341  & 0.6589 $\pm$ 0.0437  &  0.6525 $\pm$ 0.0219   \\ 
        \rowcolor{gray!15}
        w/IC-DiT    &    Synth                                           &  0.5992 $\pm$ 0.0376 & 0.6003 $\pm$ 0.0447  & 0.7924 $\pm$ 0.0175  & 0.6008 $\pm$ 0.0336  &  0.5993 $\pm$ 0.0405   \\ 
         \rowcolor{gray!15}
         w/IC-DiT     &    Real+Synth                                       & \textbf{0.6631 $\pm$ 0.0374}               & \textbf{0.6706 $\pm$ 0.0467}               & \textbf{0.8606 $\pm$ 0.0173}               & \textbf{0.6734 $\pm$ 0.0668}               & \textbf{0.6654 $\pm$ 0.0601}    \\ \midrule
        CMTA       &    Real                                & {0.6910 $\pm$ 0.0426}  & {0.6679 $\pm$ 0.0434}  & {0.8531 $\pm$ 0.0116}  & {0.6864 $\pm$ 0.0359}  & {0.6975 $\pm$ 0.0409}   \\
        w/PathDiff    &    Synth                                           & 0.6215 $\pm$ 0.0479  & 0.6046 $\pm$ 0.0319  & 0.7891 $\pm$ 0.0427  & 0.6154 $\pm$ 0.0310  & 0.6431 $\pm$ 0.0147    \\ 
        w/PathDiff    &    Real+Synth                                         &  0.6988 $\pm$ 0.0227 & 0.6785 $\pm$ 0.0401  & 0.8614 $\pm$ 0.0115  & 0.6927 $\pm$ 0.0279  &   0.7052 $\pm$ 0.0471  \\ 
        \rowcolor{gray!15}
        w/IC-DiT    &    Synth                                           &  0.6418 $\pm$ 0.0237 & 0.6351 $\pm$ 0.0317  & 0.8129 $\pm$ 0.0433  &  0.6350 $\pm$ 0.0174 &  0.6683 $\pm$ 0.0274   \\
        \rowcolor{gray!15}
        w/IC-DiT    &    Real+Synth                                           & \textbf{0.7143 $\pm$ 0.0575}  & \textbf{0.6912 $\pm$ 0.0586}  & \textbf{0.8754 $\pm$ 0.0157}  & \textbf{0.7097 $\pm$ 0.0485}  & \textbf{0.7208 $\pm$ 0.0552}    \\ \midrule
         MurreNet       &   Real                              & {0.7052 $\pm$ 0.0481}  & {0.7031 $\pm$ 0.0252}  & {0.8734 $\pm$ 0.0207}  & {0.6847 $\pm$ 0.0449}  & {0.7319 $\pm$ 0.0613}   \\
        w/PathDiff    &    Synth                                           & 0.6351 $\pm$ 0.0284  & 0.6219 $\pm$ 0.0147  &  0.7993 $\pm$ 0.0258 & 0.6143 $\pm$ 0.0372  &  0.6654 $\pm$ 0.0336   \\ 
        w/PathDiff    &    Real+Synth                                         & 0.7113 $\pm$ 0.0252  & 0.7073 $\pm$ 0.0403  & 0.8779 $\pm$ 0.0156 & 0.7004  $\pm$ 0.0273 & 0.7347 $\pm$ 0.0306   \\ 
        \rowcolor{gray!15}
        w/IC-DiT    &    Synth             & 0.6529 $\pm$ 0.0441  & 0.6618 $\pm$ 0.0348  &  0.8251 $\pm$ 0.0408   &  0.6449 $\pm$ 0.0442 &  0.6959 $\pm$ 0.0307   \\ 
        \rowcolor{gray!15}
        w/IC-DiT    &   Real+Synth                                           & \textbf{0.7255 $\pm$ 0.0316}  & \textbf{0.7154 $\pm$ 0.0392}  & \textbf{0.8811 $\pm$ 0.0420}  & \textbf{0.7251 $\pm$ 0.0728}  & \textbf{0.7415 $\pm$ 0.0448}    \\ 
         \bottomrule \bottomrule
      \end{tabular}}
      \vspace{-4mm}
   \label{tab:surv_pre}
\end{table*}

\subsection{Downstream Tasks}

\textbf{Synthetic Images on Survival Prediction.} We evaluate the utility of IC-DiT-generated synthetic pathology images for downstream survival prediction across five cancer datasets. As shown in Table \ref{tab:surv_pre}, our method demonstrates superior clinical validity under both synthetic-only and augmented training paradigms. First, when training models exclusively on synthetic data, a performance drop compared to real-data baselines is expected; however, IC-DiT incurs significantly less degradation than the state-of-the-art PathDiff synthesis method, indicating a better preservation of intrinsic prognostic features. Second, in the data augmentation setting (Real+Synth), incorporating one-fold synthetic images consistently improves the C-index scores over the real-data baselines. While both PathDiff and IC-DiT provide noticeable performance gains, integrating our IC-DiT-generated images yields substantially larger and more consistent improvements across all four baseline models and five datasets. These quantitative results confirm that our layout-controllable synthesis not only generates visually realistic images but also effectively retains clinically relevant structures and semantics, making it highly advantageous for enhancing data-efficient clinical modeling.

\noindent \textbf{Synthetic Images on Cancer Classification.} We evaluate the impact of synthetic pathology images on downstream cancer classification across five cohorts. As shown in Table \ref{tab:clas}, our IC-DiT framework demonstrates superior clinical utility under both synthetic-only and augmented training settings. In the data augmentation scenario (Real+Synth), incorporating one-fold synthetic samples consistently improves classification accuracy over the real-only baselines. While PathDiff provides moderate performance gains, augmenting with IC-DiT yields significantly larger and more consistent improvements across both baseline models (ViLa-MIL and PMIL). Notably, the most substantial improvements are observed in datasets like BLCA and BRCA, demonstrating that our layout-controllable synthesis effectively complements limited real samples.

\begin{table}[htbp]
\vspace{-4mm}
\caption{Performance comparison of the classification model on downstream tasks. Accuracy is reported across five histopathology datasets under three training settings: (1) real data only, (2) synthetic data only and (3) real data with one-fold synthetic augmentation. We highlight the best performing scores in \textbf{bold}.}
\vspace{-4mm}
\centering
\setlength{\tabcolsep}{2.5mm}
\renewcommand\arraystretch{1.2}
\scriptsize
\begin{tabular}{lccccccc}
\toprule
\toprule
\multirow{2}{*}{Methods} & \multirow{2}{*}{Train Data} & \multicolumn{5}{c}{Datasets}           \\ \cline{3-8} 
                         &                             & BLCA  & BRCA  & GBMLGG & LUAD  & UCEC & Avg. \\ \hline
ViLa-MIL                 & Real                        & 84.59 & 81.06 & 90.35  & 82.06 & 91.52 & 85.92\\
w/PathDiff               & Synth                       & 78.41 & 74.53 & 85.99  &  76.28  & 87.65 &  80.57 \\
w/PathDiff               & Real+Synth                  & 85.26 & 82.34 & 90.69  &  83.44  & 91.80 & 86.71 \\
\rowcolor{gray!15}
w/IC-DiT                 & Synth                       & 80.48 & 77.13 & 87.49  &  79.05  &  89.08 & 82.66 \\
\rowcolor{gray!15}
w/IC-DiT                 & Real+Synth                  & \textbf{88.10} & \textbf{84.73} & \textbf{91.58}  & \textbf{84.13} & \textbf{92.30} & \textbf{88.17}\\ \midrule
PMIL                     & Real                        & 87.63 & 86.90 & 91.37  & 83.41 & 92.07 & 88.28\\
w/PathDiff               & Synth                       & 80.25 & 81.34 & 85.08  & 76.47 & 86.15 & 81.86 \\
w/PathDiff               & Real+Synth                  & 88.41 & 87.54 & 91.63  & 84.96 & 92.97 & 89.10 \\
\rowcolor{gray!15}
w/IC-DiT                 & Synth                       & 82.52 & 83.79 & 88.04  & 79.44 & 88.25 & 84.41  \\
\rowcolor{gray!15}
w/IC-DiT                 & Real+Synth                  & \textbf{89.86} & \textbf{89.39} & \textbf{92.51}  & \textbf{86.26} & \textbf{94.13} & \textbf{90.43}\\ \bottomrule \bottomrule
\end{tabular}
\vspace{-8mm}
\label{tab:clas}
\end{table}

\section{Conclusions}

In this work, we have devoted to addressing the challenge of controllable high-fidelity pathology image generation by introducing In-Context Diffusion Transformers (IC-DiT), a framework that unifies textual semantics, spatial layouts, and vision-language embeddings within a diffusion transformer architecture. To overcome the impracticality of exhaustive pixel-level annotations, we have designed a multi-agent annotation pipeline that leverages large vision-language models as scalable substitutes for expert labeling, enabling semantically rich and cost-efficient supervision. Our key insight is that precise structural control for clinical utility can be achieved through cross-modal attention that aligns layout priors with generative dynamics in latent space. Extensive experiments have shown that IC-DiT surpasses existing methods in image fidelity, layout adherence, and semantic controllability, while also serving as a powerful data generation engine for downstream tasks.

\bibliographystyle{splncs04}
\bibliography{main}
\end{document}